\documentclass[10pt,twocolumn,letterpaper]{article}

\usepackage[pagenumbers]{cvpr} 

\definecolor{cvprblue}{rgb}{0.21,0.49,0.74}
\usepackage[pagebackref,breaklinks,colorlinks,allcolors=cvprblue]{hyperref}

\usepackage{amsmath}
\DeclareMathOperator*{\argmin}{\arg\!\min}
\usepackage{booktabs}
\usepackage{multirow}
\newcommand{\smallparagraph}[1]{\noindent\textbf{#1}}

\def\paperID{15018} 
\def\confName{CVPR}
\def\confYear{2025}

\title{Probing the Representational Power of Sparse Autoencoders in Vision Models}

\newcommand{\labs}{Work done at Intel Labs.} 

\author{
    Matthew Lyle Olson$^1$\thanks{\labs} \and
    Musashi Hinck$^2$\footnotemark[2] \and
    Neale Ratzlaff$^1$\footnotemark[2] \and
    Changbai Li$^3$ \and
    Phillip Howard$^4$\footnotemark[2] \and
    Vasudev Lal$^1$\footnotemark[2] \and
    Shao-Yen Tseng$^1$\footnotemark[2] \\
    $^1$Oracle \quad $^2$Intel Labs \quad $^3$Oregon State University \quad $^4$Thoughtworks \\
    {\tt\small \{matthew.olson, neale.ratzlaff, vasudev.lal, shaoyen.tseng\}@oracle.com}\\
    {\tt\small  musashi.hinck@intel.com}\\
    {\tt\small phillip.howard@thoughtworks.com \quad lc@oregonstate.edu}
}
\begin{document}

\maketitle

\begin{abstract}
Sparse Autoencoders (SAEs) have emerged as a popular tool for interpreting the hidden states of large language models (LLMs). By learning to reconstruct activations from a sparse bottleneck layer, SAEs discover interpretable features from the high-dimensional internal representations of LLMs. Despite their popularity with language models, SAEs remain understudied in the visual domain. In this work, we provide an extensive evaluation the representational power of SAEs for vision models using a broad range of image-based tasks. Our experimental results demonstrate that SAE features are semantically meaningful, improve out-of-distribution generalization, and enable controllable generation across three vision model architectures: vision embedding models, multi-modal LMMs and diffusion models. In vision embedding models, we find that learned SAE features can be used for OOD detection and provide evidence that they recover the ontological structure of the underlying model. For diffusion models, we demonstrate that SAEs enable semantic steering through text encoder manipulation and develop an automated pipeline for discovering human-interpretable attributes. Finally, we conduct exploratory experiments on multi-modal LLMs, finding evidence that SAE features reveal shared representations across vision and language modalities. Our study provides a foundation for SAE evaluation in vision models, highlighting their strong potential improving interpretability, generalization, and steerability in the visual domain. 
\end{abstract}

\setlength{\belowdisplayskip}{3pt} \setlength{\belowdisplayshortskip}{3pt}
\setlength{\abovedisplayskip}{3pt} \setlength{\abovedisplayshortskip}{3pt}

\section{Introduction}

\begin{figure}[t]
    \includegraphics[width=\linewidth]{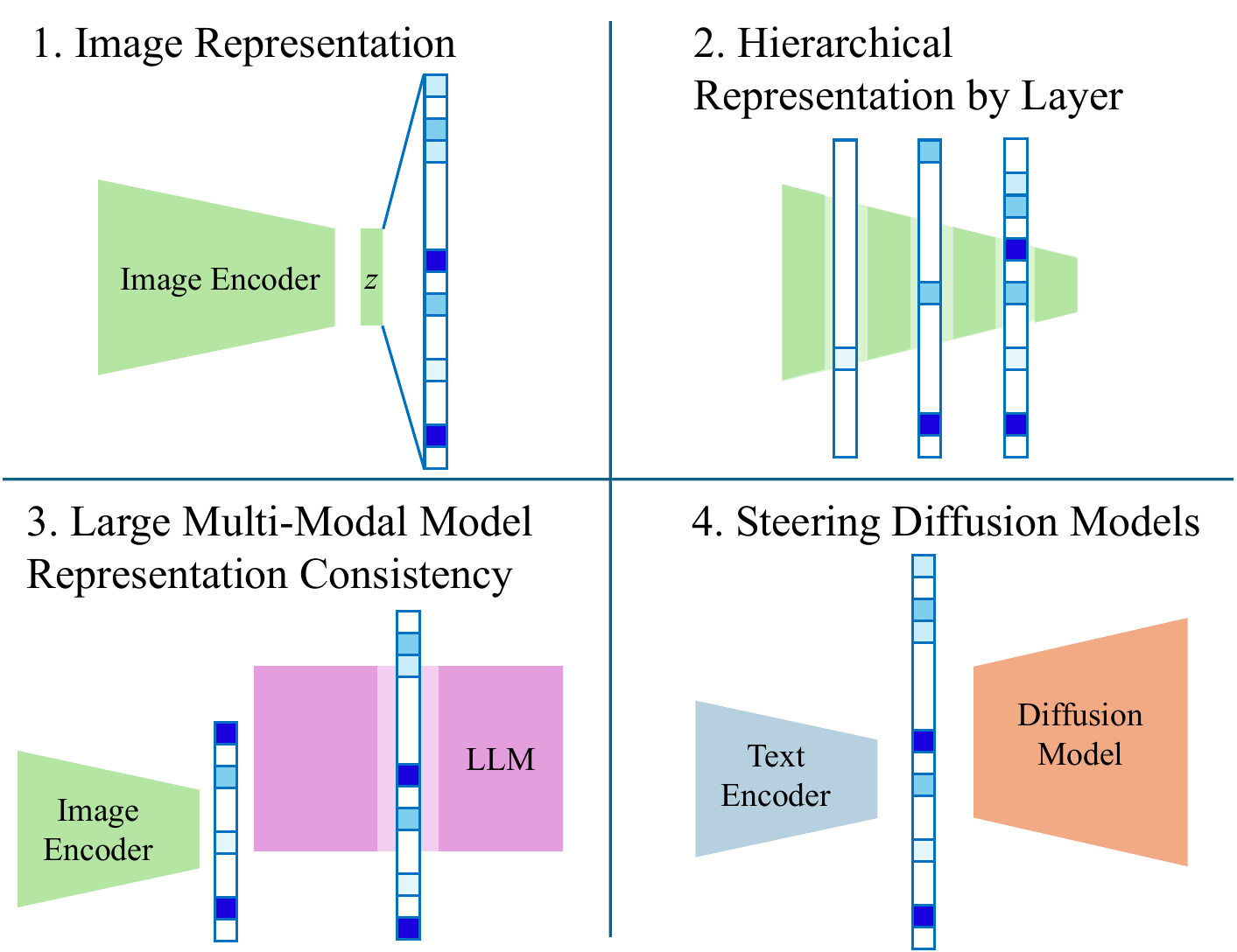}
    \vspace{-2ex}
    \caption{\textbf{An overview of the four categories of the Sparse Autoencoder (SAE) training experiments we perform}: 1) SAEs on output of vision encoders, 2) SAEs on the intermediate layers to measure hierarchical learning, 3) SAEs to understand cross modality representations in Large Multi-Modal Models, 4) Using SAEs to steer generated images in Diffusion Models. }
    \vspace{-2ex}
    \label{fig:overview}
\end{figure}

Sparse Autoencoders (SAEs) have recently gained popularity as a tool for identifying interpretable features from Large Language Models (LLMs) \cite{cunningham2023sparse,rajamanoharan2024gated_sae,gao2024scaling}.
This is done by training SAEs to reconstruct intermediary states of the LLM into an autoencoder with a large hidden layer and a sparsity penalty, which forces information compression and encourages the learning of distinct, interpretable features from the dense lower-dimensional representation space, related to the problem of dictionary learning~\cite{olshausen1997sparse}. These distinct features help alleviate the polysemanticity problem in deep neural networks, where individual neurons activate on unrelated concepts~\cite{olah2020zoom}.
These features can then be used to identify semantically meaningful ``directions'' within LLMs that can be used to steer generation \cite{turner2023activation}, or otherwise better understand \textit{why} a LLM generates particular outputs \cite{cunningham2023sparse}.
While the efficacy of SAEs in the text domain has been explored in numerous studies~\cite{minegishi2025rethinking,kantamneni2025sparse,mudide2024efficient}, there is still little work extending their application to visual modalities.
In this work, we address this gap by providing an extensive evaluation of SAEs for the visual domain and demonstrate that their utility as an interpretability and steering tool extends to the visual domain.

Based on analogous applications of SAEs to LLMs, we propose a collection of tasks and evaluation metrics for probing the utility of SAE features in vision models.
Our analysis, summarized in Figure~\ref{fig:overview}, extends SAEs to multiple types of deep vision models, from image encoders to Large Multi-modal Models (LMMs). 
For each analyzed model, we find evidence that suggests SAEs extract semantically meaningful features, which can be used for downstream tasks including out-of-distribution (OOD) generalization and steering.
For image encoders, we evaluate SAE features for two applications: OOD generalization and ontology construction.
We fit our SAEs on the DINOv2~\cite{oquab2023dinov2} visual foundation model and ImageNet Dataset~\cite{russakovsky2015imagenet}, and find that the extracted features achieve competitive performance on OOD generalization.
For LMMs, we conduct exploratory experiments on LLaVA 1.5 \citep{liu2024visual} to investigate how SAEs reveal vision-language representation alignment. We observe qualitative evidence of feature overlap between image and text modalities when processing the same concepts, suggesting shared internal representations. Additionally, our analysis indicates that visual features persist through early LLM layers rather than immediately collapsing, though we note these findings are preliminary and would benefit from future quantitative validation.
Finally, we evaluate whether SAEs can be used to steer image generations for diffusion models such as Stable Diffusion \citep{rombach2021highresolution}. Despite the challenges posed by significant differences in model architecture between diffusion models and transformers, we show that features discovered by SAEs at generation time can be used to robustly steer the output of a diffusion model.

Taken together, our work pushes the frontier on multiple fronts in the field of interpretability methods for deep vision models. We demonstrate how to introduce SAEs to three architectures (transformer-based image encoders, LMMs, and diffusion models) and provide evidence of their efficacy for a range of applications. In sum, our contributions are as follows:
\begin{enumerate}
    \item We extend SAEs to a myriad of modern vision architectures such as encoders (DINOv2), diffusion models (Stable Diffusion), and LMMs (LLaVA). 
    \item We develop novel evaluations and metrics for SAEs in the vision domain, and show that they yield valid and interpretable features.
    \item We use these features for several downstream applications, including diffusion model steering and competitive performance on OOD generalization.
\end{enumerate}

\section{Background}
\subsection{Sparse Autoencoders}

We study three SAE variants: ReLU~\cite{cunningham2023sparse}, TopK~\cite{gao2024scaling}, and Gated~\cite{rajamanoharan2024gated_sae}. These provide different sparsity-inducing mechanisms and have been explored in prior work.

\paragraph{Baseline: ReLU SAEs.} The ReLU SAE follows a straight forward setup. Given an input vector $x \in \mathbb{R}^d$ from the model representation space, the encoder and decoder are defined as:
\begin{align}
    z &= \text{ReLU}(W_{\text{enc}}x  + b_{\text{enc}}) \\
    \hat{x} &= W_{\text{dec}}z + b_{\text{dec}}
\end{align}
where $W_{\text{enc}} \in \mathbb{R}^{n \times d}$, $b_{\text{enc}} \in \mathbb{R}^n$, $W_{\text{dec}} \in \mathbb{R}^{d \times n}$, and $b_{\text{dec}} \in \mathbb{R}^d$. The loss function consists of reconstruction error and an L1 sparsity penalty:
\begin{equation}
    L = \|x - \hat{x}\|^2_2 + \lambda \|z\|_1.
\end{equation}

\paragraph{TopK SAEs.} The TopK SAE modifies the encoder activation function by keeping only the $k$ largest latents and setting the rest to zero:
\begin{equation}
    z = \text{TopK}(W_{\text{enc}}x).
\end{equation}
This allows explicit control over the number of active features per input.

\paragraph{Gated SAEs.} The Gated SAEs introduce a gating mechanism to separate feature selection from magnitude estimation. The encoder is defined as:
\begin{equation}
    \begin{aligned}
       \tilde{f}(x) = &\mathbf{1}[(W_{\text{gate}}(x - b_{\text{dec}}) + b_{\text{gate}}) > 0] \\
        &\odot \text{ReLU}(W_{\text{mag}}(x - b_{\text{dec}}) + b_{\text{mag}})
    \end{aligned}
\end{equation}
where where $\mathbf{1}[\boldsymbol{\cdot} > 0]$ is the (pointwise) Heaviside step function and $\odot$ denotes elementwise multiplication.
A gating loss encourages sparsity while ensuring selected features contribute meaningfully to reconstruction.

Each of these SAE variants provides different trade-offs in interpretability and sparsity, which we explore across multiple vision tasks.

\subsection{Related Work}
\paragraph{Sparse Autoencoders in Vision}
The study of SAEs is a rapidly growing field, and recent work has been done towards their applications in vision topics.
\citet{rao2024discover} train SAEs on CLIP for concept bottleneck models, achieving interpretable classification on standard benchmarks. Our work extends this by examining OOD generalization capabilities.
\citet{stevens2025sparse} studies SAEs trained on the patch embeddings of image encoder models, whereas we focus on a wide scale study of class embeddings and multi-modal models.
Concurrent work is also exploring SAE-based diffusion steering~\cite{daujotas2024interpreting, surkov2024unpacking, kim2025concept,cywinski2025saeuron,tinaz2025emergence} by using different architectural approaches (e.g., U-Net integration). Our text-encoder approach provides a complementary method that operates at the semantic level.

\paragraph{Interpretability Tools for Vision Models}
\citet{gandelsman2023interpreting} propose an interpretability framework for CLIP-like models based on decomposing image features as a linear combination of individual image patches, model layers, and attention heads.  Similar approaches have been proposed for extending this decomposition-based interpretability framework to other Vision Transformer (ViT) variants beyond CLIP \citep{balasubramanian2025decomposing}. In the realm of LMM interpretability, \citet{parekh2024concept} propose a dictionary learning approach to extract concepts that help explain internal model representations. \citet{stan2024lvlm} introduce an interpretability tool for LMMs named LVLM-Interpret which visualizes how model outputs relate to input images through relevancy maps and attention values. \citet{yu2024understanding} also develop an interpretability tool for LMMs which is based on mechanistic interpretability methods. Comparatively less attention has been paid to interpretability methods for text-to-image diffusion models. \citet{dombrowski2024trade} demonstrate a trade-off between the performance of diffusion models and their ability to be interpreted. Methods for learning semantically meaningful directions in the latent space of diffusion models have also been proposed \citep{haas2024discovering}.
In contrast to these prior works, our study focuses on applying a single interpretability approach (SAEs) to multiple different types of vision models for improving model interpretability, OOD generalization, and controllability via steering.

\section{Image Encoders}


Also known as visual foundation models, image encoders such as CLIP~\cite{radford2021clip} are used to extract dense representations of visual inputs for other downstream tasks such as zero-shot image classification, object detection and captioning.
For our experiments we use the state-of-the-art DINOv2~\cite{oquab2023dinov2}, which also uses a ViT architecture \cite{dosovitskiy2021an}, but with an ensemble of training objectives.
We assess the suitability of SAEs to image encoders through two evaluations by analyzing whether the learned features demonstrate generalization and by latent feature discovery.

Our first evaluation tests the generalization of these SAE features through domain shift experiments. We train linear classifiers on SAE features extracted from ImageNet train samples, then test classification accuracy on shifted domains with identical label sets: ImageNet-V2~\cite{imagenetv2} (new photos), ImageNet-Sketch~\cite{imagenet_sketch_dataset} (pencil sketches), ImageNet-A~\cite{imagenet_a_dataset} (adversarial natural images), and ImageNet-R~\cite{imagenet_r_dataset} (artistic renditions). For feature reconstruction analysis, we leverage the OpenOOD framework~\cite{zhang2023openood} to evaluate how well different SAEs can autoencode OOD data of varying similarity to ImageNet. 

For our second evaluation, we test whether SAEs capture \textit{ontological features} in vision encoders by identifying SAE features that map to higher-order concepts. The ImageNet classes are drawn from the WordNet ontology, which relates these classes as a hierarchical tree of \textit{synsets}. We identify SAE features that activate on groups of ImageNet classes that belong to the same higher-level WordNet concept.

We organize this section as follows: in the first part, we provide the evaluation datasets and metrics. In the second part, we explain the ways in which we fit SAEs to image encoders. In the final part, we provide experimental results.
Our results show that SAEs provide valid features that generalize to OOD settings and capture complex semantic structures within image encoder models.

\subsection{Datasets}

\smallparagraph{Training Data.} We train SAEs on ImageNet \cite{imagenet_dataset} training set (1.2M images, 1000 classes) using the class token embeddings from pretrained vision encoders.

\smallparagraph{OOD Generalization Benchmarks.} We evaluate generalization using datasets with identical label spaces to ImageNet, but different visual domains:
\begin{itemize}
\item \textit{ImageNet-V2} \cite{imagenetv2}: 10K images from Flickr with matched class distribution (tests robustness to minor distribution shift)
\item \textit{ImageNet-S} \cite{imagenet_sketch_dataset}: 50K pencil sketches (tests shape bias vs texture bias)
\item \textit{ImageNet-A} \cite{imagenet_a_dataset}: 7.5K adversarially filtered natural images (tests robustness to hard examples)
\item \textit{ImageNet-R} \cite{imagenet_r_dataset}: 30K artistic renditions including paintings, cartoons, and sculptures (tests abstraction)
\end{itemize}

\smallparagraph{OOD Detection Benchmarks.} For reconstruction analysis, we use the OpenOOD framework \cite{zhang2023openood} with near-OOD (SSB-Hard~\cite{ssb_hard_dataset}, NINCO~\cite{ninco_dataset}) to measure feature reconstruction on datasets that are semantically similar to ImageNet, and we use far-OOD (iNaturalist~\cite{iNaturalist_dataset}, Textures~\cite{textures_dataset}, OpenImage-O~\cite{open_image_o_dataset}) datasets to measure feature reconstruction of dissimilar data.
\begin{figure*}[tb]
    \includegraphics[width=\linewidth]{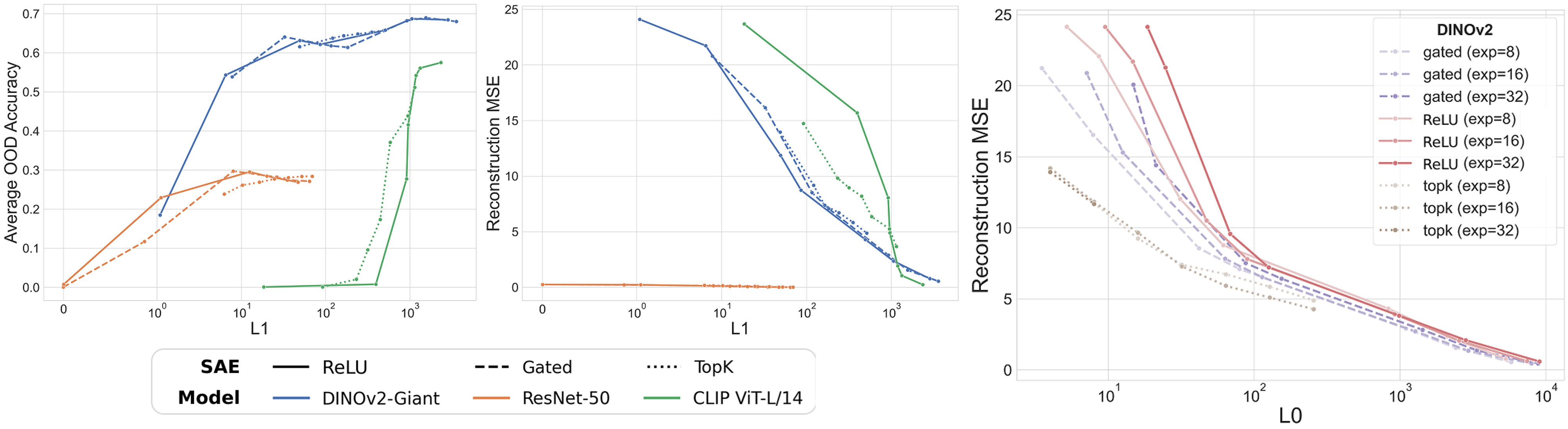}
    \vspace{-2ex}
    \caption{
    \textbf{Measuring the generalization of various SAEs and encoder models on OOD data.}
    \textbf{(Left):} The performance of SAE representations when used to train a linear classifier across different $\lambda$ penalties. We find that DINOv2 is robust to a range of SAE hyperparmeters. In contrast, CLIP performance suffers at high levels of sparsity.
    \textbf{(Middle):} The average activation magnitude compared to average OOD reconstruction across SAEs and models. We find that the trained ResNet features are highly specific to ImageNet, trivializing the autoencoding task. In contrast, DINOv2 and CLIP contain more general features, leading to an initially more difficult training regime. 
    \textbf{(Right):} SAE metrics with repsect to DINOv2 across a range of training configurations. We measure OOD reconstruction versus sparse activations, and we find that reconstruction error decreases aimilarly across SAE types and hidden expansion sizes.
    }
    \label{fig:mse_results}
\end{figure*}

\subsection{Evaluation Metrics}

\paragraph{MSE Reconstruction Error.} The reconstruction quality is evaluated using the mean squared error (MSE) between the input $x$ and the reconstructed output $\hat{x}$:
\begin{equation}
    \text{MSE} = \frac{1}{d} \|x - \hat{x}\|_2^2
\end{equation}


\paragraph{L1 Sparsity.} The L1 norm of the latent representation $z$ quantifies the level of sparsity after training:
\begin{equation}
    \|z\|_1 = \sum_{i=1}^{n} |z_i|
\end{equation}

\paragraph{$L_0$ Activation Count.} The $L_0$ norm measures the number of active (nonzero) latent units:
\begin{equation}
    \|z\|_0 = \sum_{i=1}^{n} \mathbf{1}(z_i \neq 0)
\end{equation}
where $\mathbf{1}(\cdot)$ is an indicator function. This metric directly quantifies the sparsity level by counting active units.

For the second evaluation, we leverage the hierarchical structure of the ImageNet classes.
The ImageNet-1k dataset contains $1000$ classes, which are synsets in the WordNet ontology.
The parents of a synset are \textit{hypernyms}, and its children are \textit{hyponyms}.
\footnote{
    e.g., \textit{Pembroke Corgi} is an ImageNet class, which is an hyponym of \textit{Corgi}, which is in turn a hyponym of \textit{Dog}. \textit{Dog} is a hypernym of \textit{Corgi} and \textit{Pembroke Corgi}, and \textit{Corgi} is a hypernym of \textit{Pembroke Corgi}.
}

We are interested not only in SAE features that activate on the $1000$ leaf-level classes, but also the extent to which SAE features that activate on multiple classes are capturing higher levels in the ImageNet concept hierarchy. In order to measure this we construct two metrics, which we call \textit{LCH Height} and \textit{Ontological Coverage}.

Let $\Omega$ be the set of leaf ImageNet classes ($|\Omega|=1000$ for ImageNet-1k), and let  $\mathcal{S}$ be the set of all WordNet synsets that occur as ancestors (including self) of any leaf in $\Omega$. Thus $\Omega \subset \mathcal{S}$, but $\mathcal{S}$ is not the set of all WordNet synsets. For each synset $h \in \mathcal{S}$, we denote its leaf set as:
\begin{equation}
    L(h) = \{\omega \in \Omega: \text{there is a hypernym path }\omega\text{ to }h\}
\end{equation}

Thus, given the synset $h$, $L(h)$ is the set of all ImageNet leaf classes that are descendants (hyponyms) of $h$. Note also that for all $\omega \in \Omega$, $L({\omega})=\{\emptyset\}$.

We denote the set of classes an SAE feature activates on as $C_k \subseteq \Omega$.
For a given set $C_k$, the \textit{lowest common hypernym} $h_k$ is the synset with the smallest subtree that contains all elements of $C_k$. This is analogous to \textit{lowest common ancestor}.
\begin{equation}
    h_k = \text{LCH}(C_k) = \argmin_{h \in \mathcal{S}:\;C_k \subseteq L_h}|L({h_k})|
\end{equation}

\paragraph{LCH Height} of $C_k$ is calculated the average height of $h_k$. This is equivalent to the average path distance between the elements of $C_k$ and $h_k$:
\begin{equation}
    \text{LCH Height}(C_k) = \frac{1}{|C_k|} \sum_{\omega \in C_k}\text{dist}(\omega, h_k)
\end{equation}

\paragraph{Ontological Coverage} is calculated as the proportion of elements in $C_k$ that are in $S_{h_k}$:
\begin{equation}
    \text{Coverage}(C_k) = \frac{|C_k|}{|L(h_k)|}
\end{equation}

These two metrics measure how well an SAE feature captures a higher-order class in the ImageNet hierarchy. LCH height indicates the relevant level of abstraction that an SAE feature may be capturing, and ontological coverage indicates how well it captures a higher-order concept.
There are some limitations to this metric: SAE features activating on a single ImageNet class will have a coverage of $1$, and the use of the LCH as the relevant set may overly penalize SAE features that are largely coherent except for a single element. We thus consider the two metrics in tandem.

\subsection{Implementation}

We conduct an extensive hyperparameter sweep across the sparsity regularization coefficient $\lambda \in$\texttt{ [0.1, 0.5, 1.0, 5.0, 10.0, 50.0 100.0]} or $k \in$ \texttt{[4, 8, 16, 32, 64, 128, 256]}, we use three expansion factors \texttt{[8, 16, 32]}, three SAE architectures: ReLU, TopK, and Gated-- and three base image encoder models: DINOv2~\cite{oquab2023dinov2}, CLIP~\cite{radford2021clip}, and ResNet~\cite{he2016deep}.

The input to the SAEs are the class embedding output by the base image encoder at a selected layer. All OOD-generalization experiments use the final layer's output.
All SAEs are trained for three epochs and minibatch of size $64$, with an Adam\cite{kingma2014adam} optimizer using a $1e^{-4}$ learning rate with  5\% linear warm-up and 20\% linear decay. We also use a 5\% $\lambda$ warm-up to minimize dead neurons. We use the SAELens library~\cite{bloom2024saetrainingcodebase} for our training.

All images are resized to $224$ for fair comparison. Linear models fit on any embeddings (SAE or original) are fit with identical training hyper-parameters.

\subsection{Feature Generalization Results}

Figure \ref{fig:mse_results} shows
high level metrics for all our training runs. We report how OOD accuracy and reconstruction changes with sparsity constraints. We find that the DINOv2 model outperforms the other foundation models. We also find the trade-off between reconstruction and sparsity is consistent with language-only SAE trends~\cite{gao2024scaling}. We also find TopK SAE to do well on OOD-reconstruction, but little difference in SAE architecture for classification. We analyze these metrics in more detail in Supplement Section \ref{sec:sae_results}.

\begin{table}[t]
    \centering
    \renewcommand{\arraystretch}{1.2}
    \begin{tabular}{l c c c c c }
        \toprule
        \multirow{2}{*}{Method}  & \multicolumn{1}{c}{ } & \multicolumn{4}{c}{ImageNet OOD} \\
          \cmidrule(lr){3-6}
                &  Val & V2 & S & A & R  \\
        \midrule
        LoRA \cite{hu2022lora}                    & 72.68 & 65.57 & 48.61 & 49.39 & 76.29 \\
        WiSE-FT \cite{wortsman2022robust}               & 73.91 & 66.69 & 49.67 & 49.67 & 77.11 \\
        LP-FT \cite{kumar2022fine}                   & 71.65 & 62.69 & 43.35 & 40.49 & 69.77 \\
        FLYP \cite{goyal2023finetune}                    & 73.26 & 65.39 & 49.21 & 49.40 & 77.13 \\
        
        \midrule
        DINOv2       & 83.23 & 75.00 & 62.18 & 58.19 & 75.15  \\
        DINOv2+SAE   & \textbf{83.84} & \textbf{75.46} & \textbf{63.38} & \textbf{61.83} & 66.31  \\
        CLIP         & 79.67 & 70.16 & 53.50 & 42.07 & 57.54  \\
        CLIP+SAE     & 81.85 & 72.54 & 53.83 & 39.56 & 55.77  \\
        ResNet       & 80.14 & 68.61 & 28.75 & 05.00 & 27.34  \\
        ResNet+SAE   & 78.28 & 66.38 & 25.37 & 02.61 & 23.21  \\
        \bottomrule
    \end{tabular}
    \caption{\textbf{Comparison of different foundation model fine-tuning methods on ImageNet and out-of-distribution (OOD) datasets}. Top half are baselines and bottom half are linear layers fit to vision encoder embeddings.
    We report values for $\lambda=0.5$, and 8-fold expansion for the SAEs. We find fitting a linear layer to the SAE's representation to often improve accuracies across the board.}

    \label{tab:results_small}
\end{table}

In Table \ref{tab:results_small}, we show the results of OOD-generalization accuracy, where baselines are various ways of fine-tuning a pretrained models: added parameters (LORA~\cite{hu2022lora}), linear-interpolation between the original and finetuned model (WiSEFT~\cite{wortsman2022robust}), fitting a linear classifier before finetuning the original model (LP-FT~\cite{kumar2022fine}), and using text labels (FLYP~\cite{goyal2023finetune}). 

we show the striking results that SAEs do better than original embedding baselines on OOD-generalization. We also note the ResNet's SAE is worse than the baseline. We believe this is from the non-generalized nature of it's latent space, as it has only been trained on ImageNet. We also speculate that by training the SAEs on ImageNet before fitting a linear layer, we enhance the ease of fitting the class predictions \footnote{While the predictive classifier heads we trained are all consistently trained, we hypothesize extensive hyper-parameter tuning may increase baseline's score (which we were unable to do for computational reasons)}. Finally, as an ablation, we trained SAEs with $\lambda$=0 and found similar results the baseline embeddings.

\subsection{Ontological Feature Results}

\begin{figure}
    \centering
    \includegraphics[width=1.0\linewidth]{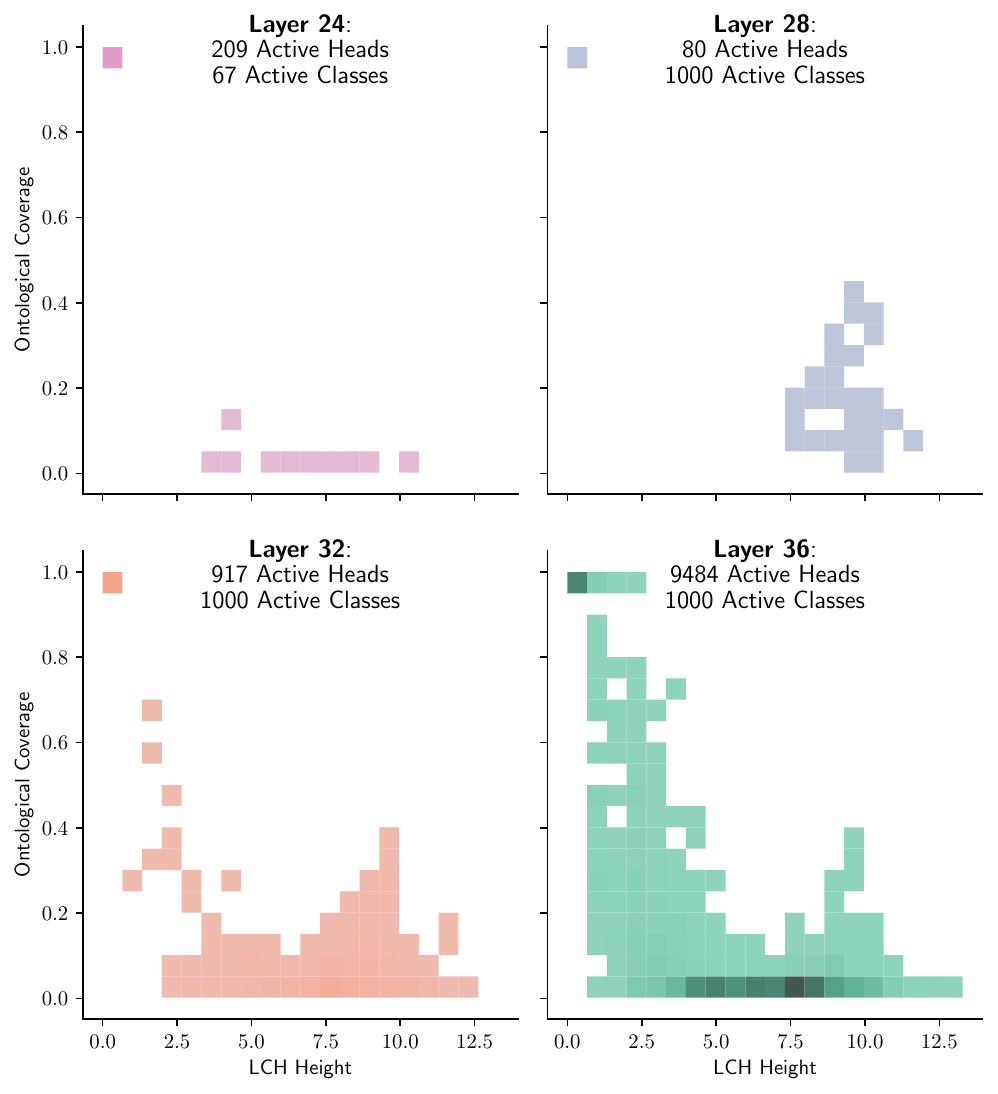}
    \caption{
    \textbf{Distribution of LCH Height vs Ontological Coverage for SAE features at Layer 24, 28, 32 and 36 of DINOv2.}
    For each layer, we plot the distribution of LCH height and ontological coverage of the SAE features. Darker indicates higher bin density.
    }
    \label{fig:lch_coverage}
\end{figure}

Figure \ref{fig:lch_coverage} shows the distribution of coverage and LCH height for SAE features from layer 24, 28, 32, and 36 of DINOv2. We train an SAE on every layer, finding  there is little SAE feature activation in earlier layers the model (we explore all these layer-wise SAEs in Supplement Section \ref{sec:dino-by-layer}).

The SAE features at layer 24 largely either activate on a single class (top-left corner of top-left subplot) or on a large number of dissimilar classes (strip along bottom).
Later layers have increasing numbers of SAE features with high ontological coverage. In particular, layer 36 has 90 multi-class SAE features with ontological coverage of $1.0$, capturing higher-order groups of things such as \textit{elasmobranch} (sharks and rays), \textit{whales}, \textit{woodwind instruments} and \textit{warships}.

\begin{table}[tb]
\centering
\begin{tabular}{lccc}
\toprule
Ontological Coverage &  $>0.99$ &  $>0.75$ & \# Features \\
\midrule
Original encoder & 0 & 0 & 61,440 \\
Random vectors & 6 & 6 & 491,520 \\
\textbf{SAE features} & \textbf{479} & \textbf{593} & 491,520 \\
\bottomrule
\end{tabular}
\caption{Counts of latent features above different Ontological Coverage thresholds for all layers on DINOv2. We find SAEs learn sparse features that capture superordinate concepts compared to a random vector baseline and to the original DINOv2 features. }
\label{tab:ontology_baseline}
\end{table}

We computed the Ontological Coverage score on all 40 DINOv2 layers in Table~\ref{tab:ontology_baseline}, and find SAE features capture high coverage features at a substantially higher than random projections or baseline features.


\begin{figure}
    \centering
    \includegraphics[width=0.9\linewidth]{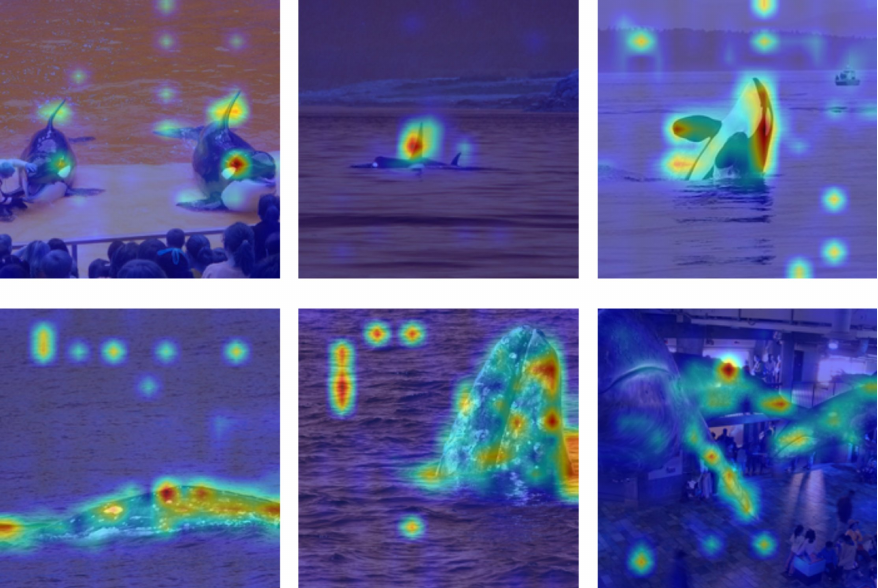}
    \caption{
    Relevancy maps of the hierarchical SAE feature at DINOv2 Layer 36 activating on images of whales. These relevancy maps show the model highly activating on the hierarchical concept of both Orcas and Grey Whales, which show DINOv2's ability to focus on highly meaningful parts of an image.
    }
    \label{fig:xai_rm}
\end{figure}

We further assess the spatial alignment of SAE features by visualizing the relevancy maps \cite{chefer2021generic} of the feature activations, as shown in Figure \ref{fig:xai_rm}. Given an image $I$, we generate feature-wise heatmaps highlighting important regions responsible for the activation of each sparse feature, providing insight into the grounding of interpretable features.

\section{Diffusion Model Steering}

In this section, we describe our experiments applying sparse autoencoders to diffusion models for image generation. 
The goal of these experiments is to analyze whether specific SAE features correspond to semantically meaningful attributes in text prompts and whether manipulating these features during inference results in consistent semantic changes in the generated images.

\begin{figure*}
\includegraphics[width=\linewidth]{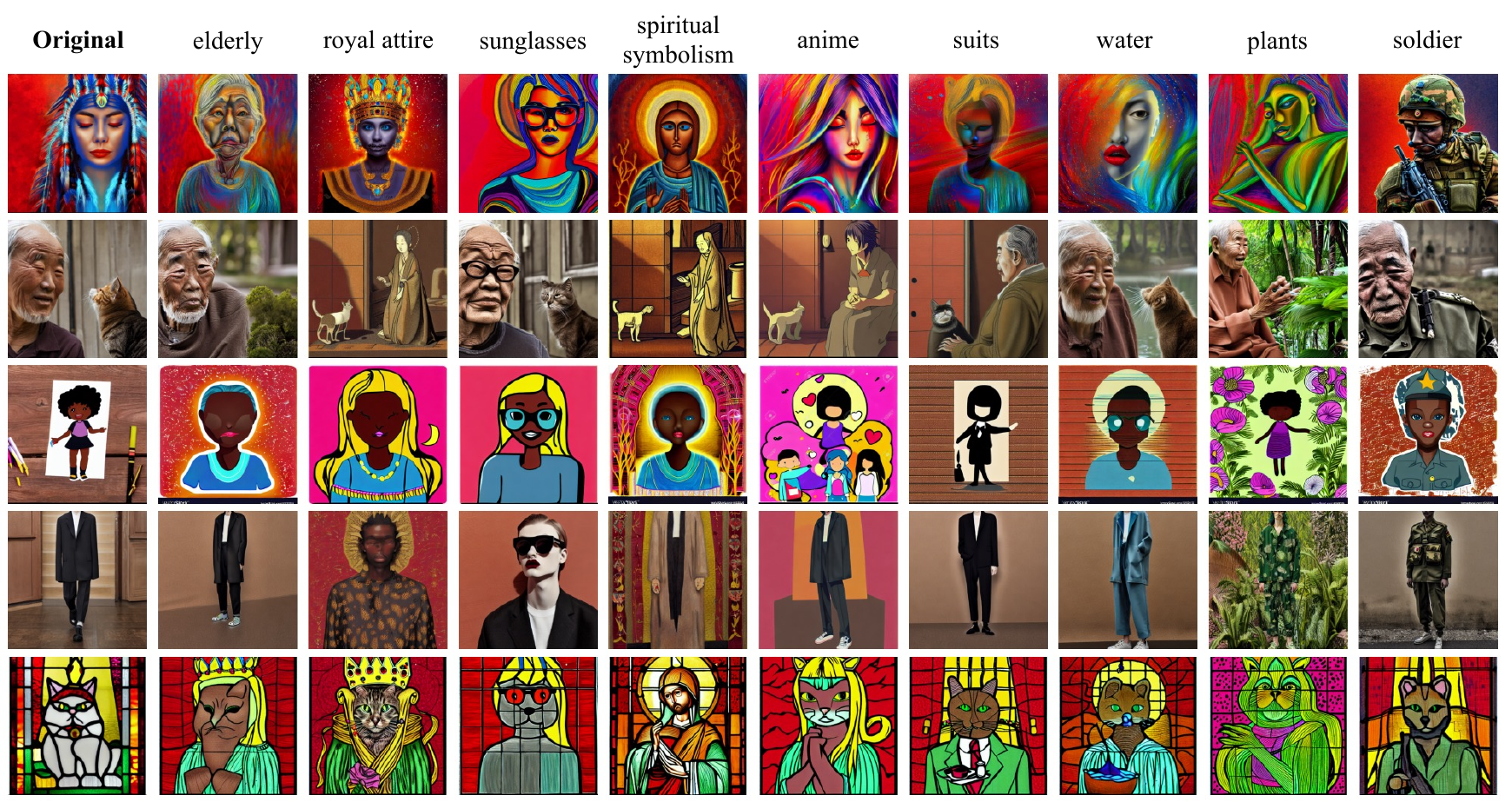}
\caption{
\textbf{Example images from steering Stable Diffusion v1.4.} 
The original generated images without steering are shown in the first column and the following columns are the results of steering in the directions of different SAE features.
Attributes such as \textit{elderly}, \textit{sunglasses}, \textit{spiritual symbolism}, and \textit{anime} emerge as steering is applied by each feature.
The attribute labels are identified through prompting a large multimodal model on a separate set of images.
}
\label{fig:diffusion_steer}
\end{figure*}

\subsection{Steering Diffusion Models using SAE}

We train a sparse autoencoder on the penultimate layer of the text encoder from Stable Diffusion \cite{rombach2021highresolution}, treating each text token’s embedding as an independent data sample. Given a large corpus of text prompts, we extract their token representations from this layer and fit an SAE with the objective of learning sparse, disentangled features. The resulting SAE consists of multiple features, each capturing a different dimension of variation in the text representations.

To steer diffusion models during image generation, we manipulate specific SAE features to adjust their influence on the final output. 
Specifically, for a given text prompt, we apply the trained SAE to the text encoder output and amplify the text embeddings in the direction of the target SAE feature with a scaling coefficient $\lambda$, as well as use Classifier Free Guidance \cite{ho2022classifier} to steer the negative embeddings away from the target feature.
Through this, the text encoder output is steered towards the semantic attribute of the SAE feature before being passed to the denoising model.

\subsection{Mapping SAE Features to Semantic Attributes}
We design a systematic analysis to determine if individual SAE features correspond to human-interpretable attributes.

First, we extract token-wise activations of the trained SAE features using captions from MS-COCO \cite{lin2014microsoft}.
After mapping the activations back to text within the captions, we then filter for descriptive words and only select features that activate for nouns, adjectives, and adverbs.
Through this method, we obtain a subset of relevant features that represented around 10\% of the total number of SAE features. 

Next, we generate sample images using prompts from Midjourney Prompts \cite{midjourneyPrompts} while also applying steering using each of the SAE features identified in the first stage. 
Using the generated original and steered images, we prompt an LMM to identify attribute differences between the two sets of images. 
In our experiments we used 4 sample images and Qwen2.5-VL \cite{bai2025qwen25vltechnicalreport} as the LMM to return the most important keywords that differentiate between the original and steered images for each feature.
We then select the most dominant keyword for each feature as the main attribute based on inverse document frequency across all keywords. 
We show randomly sampled examples of identified feature attributes:
\begin{center}
\begin{minipage}{0.95\linewidth}
    \hrule\vspace{4pt}  
    \textit{bird, desert, neon, abstract art, cartoon character, blonde hair, sea, elderly, creative expression}
    \vspace{4pt}\hrule  
\end{minipage}
\end{center}

\paragraph{Steering on Semantic Attributes} Examples of generated images with steering are shown in Figure \ref{fig:diffusion_steer}.
We sample prompts from Midjourney Prompts and steered the diffusion model in the direction of specific SAE features during image generation. 
The semantic attributes of each SAE feature as obtained from the LMM are displayed above the steered images. 
With these examples, we observe that the generated images do indeed exhibit the properties of the attribute labels of the features after steering. 

In supplement Section \ref{sec:diffusion35} we show qualitative steering examples for StableDiffusion3.5~\cite{stabilityai2024sd35}. We found this model substantially harder to steer due to its complex text encoding architecture with three separate text embedding models.

\paragraph{Attribute Validation with CLIP}

To assess the semantic consistency of steered images with respect to SAE feature attributes, we compute CLIP-based image-text similarity scores.  
Let $f_I$ and $f_T$ be the CLIP image and text encoders. Given an original image $I_\text{orig}$, a steered image $I_\text{steer}$, and a text prompt $t_\text{orig}$, we compute their embeddings:  
$$
z_\text{orig} = f_I(I_\text{orig}), \quad z_\text{steer} = f_I(I_\text{steer}), \quad z_t = f_T(t_\text{orig}).
$$  

We also define an extended prompt incorporating the target attribute, $t_{\text{attr}} = t_\text{orig} + \text{attr}$, with embedding $z_{\text{attr}} = f_T(t_{\text{attr}})$.  
We can then measure the cosine similarity between these embeddings to determine semantic similarity.

From Table \ref{tab:diffusion_clip}'s results, we observe that steered images have higher similarity to the input prompt that has the attribute keyword of the steering feature appended. 
The results validate our steering hypothesis with high statistical significance ($p < 0.0001$, paired t-test, $n=100$ prompt-attribute pairs). The 5.8\% increase in similarity to attribute-augmented prompts demonstrates that SAE features capture semantically meaningful directions for generation control.

\begin{table}[tb]
    \centering
    \begin{tabular}{c c c}
     \toprule
     \textbf{Image}  & \textbf{Text}  & \textbf{Cosine similarity} \\
     \midrule     
     $I_\text{orig}$ & $t_\text{orig}$   &  0.264 \\
     $I_\text{steer}$ & $t_\text{orig}$    &  0.216 \\
     $I_\text{steer}$ & $t_{\text{attr}}$    &  0.228 \\
     \bottomrule
    \end{tabular}
    \caption{CLIP cosine similarities of generated images to different text configurations. We find the image generated by steering is more similar to a representation with the target text attribute added to the caption.}
    \vspace{-2ex}
    \label{tab:diffusion_clip}
\end{table}

\section{Large Multi-Modal Models}

We extend our analysis to Large Multi-modal Models (LMMs) through exploratory experiments on LLaVA 1.5 \cite{liu2024visual}. While our image encoder and diffusion model experiments provide rigorous quantitative validation, our LMM investigation offers preliminary qualitative insights into how SAEs reveal vision-language representation alignment.

LMMs must bridge visual and textual representation spaces to produce coherent outputs. We investigate whether SAEs can illuminate this bridging by examining: (1) representational overlap between image and text processing for the same concepts, and (2) how visual representations evolve through the language model layers. We use LLaVA due to its straightforward architecture where image and text tokens are concatenated.

\subsection{Multimodal Representation Overlap}

\begin{figure}
\includegraphics[width=\linewidth]{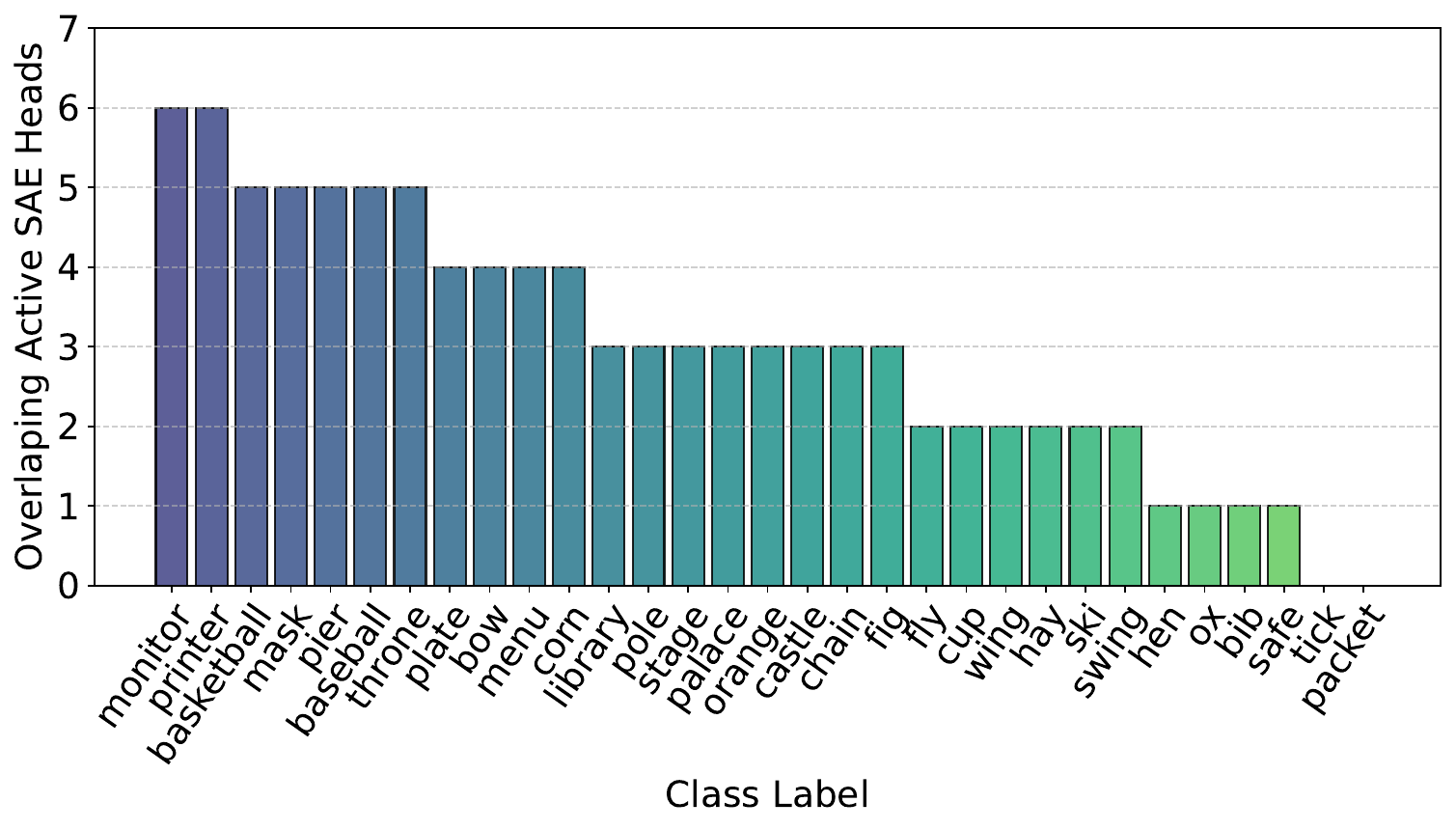}
\vspace{-1ex}
\caption{\textbf{Representational Overlap of LLaVA Text and Image Features.} Top 1\% of activated SAE features for both image and text representations of the same ImageNet class. Despite high sparsity, we observe consistent feature overlap across modalities for most examined classes.}
\vspace{-2ex}
\label{fig:llava_image_text_sae}
\end{figure}

We explore whether LLaVA uses shared representations across modalities by training a ReLU SAE on middle-layer activations during ImageNet captioning. Following prior work \citep{turner2023activation, mini2023understanding, subramani2022extracting}, we focus on middle layers where representations are most abstract. We then conduct a two-stage comparison:

\smallparagraph{Text Representations.} We prompt LLaVA (without images) to define a concept (e.g., "What is a clock?") and extract SAE features from the resulting activations.

\smallparagraph{Image Representations.} We input an image from the same class and request a description (without mentioning the class label), then extract SAE features from LLaVA's response.

Figure~\ref{fig:llava_image_text_sae} shows the large overlap in activated SAE features between modalities. For nearly all examined classes, similar SAE features activate whether LLaVA processes textual or visual information about that concept. This pattern suggests LLaVA may map both modalities to a shared representational space, though we acknowledge this remains a qualitative observation requiring future validation.

\subsection{Layer-Wise Visual Representation Evolution}

\begin{figure}
\includegraphics[width=\linewidth]{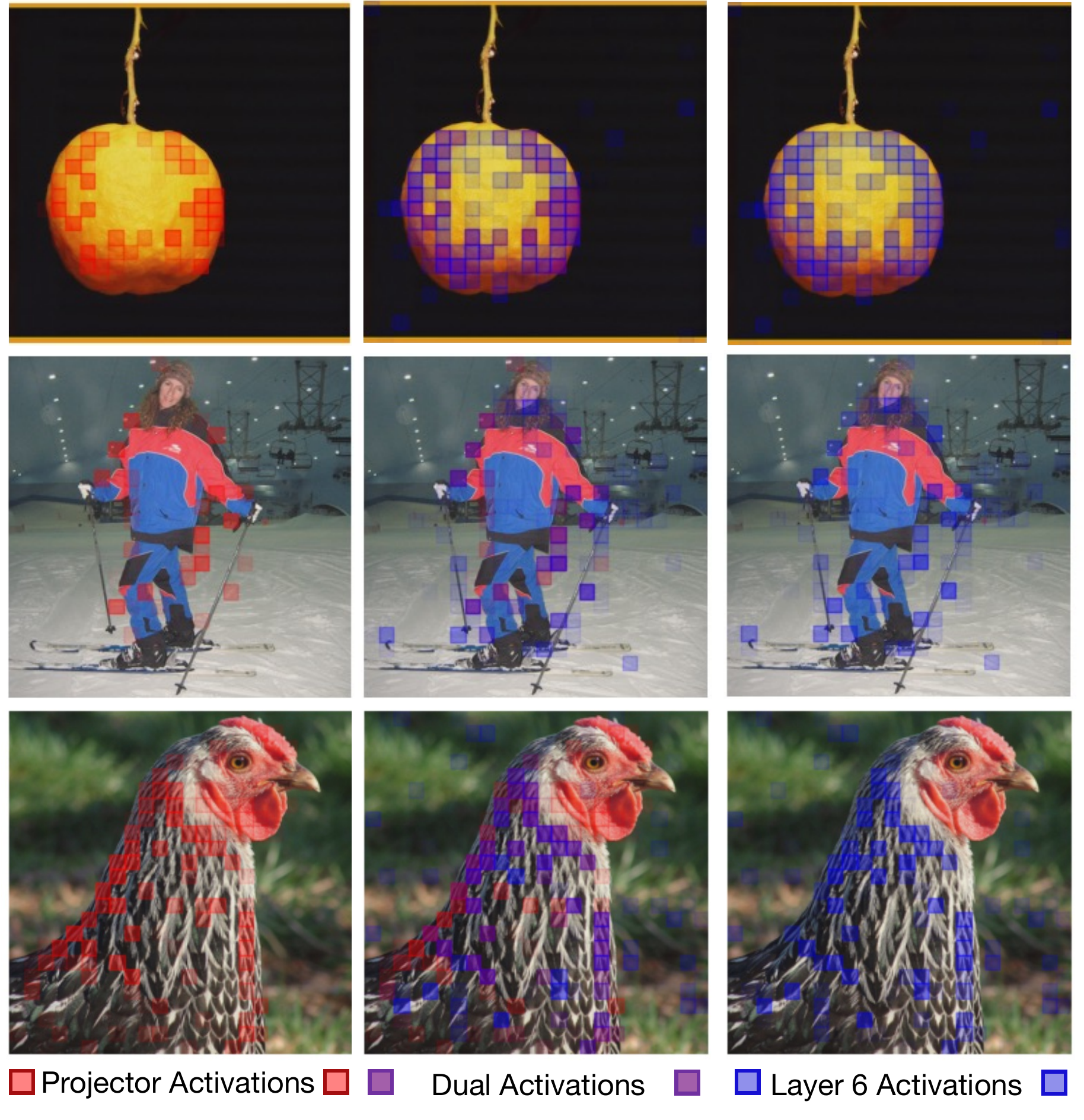}
\caption{\textbf{Evolution of visual representations through LLaVA.} Activated image patches from SAE trained on projector output (left) versus layer 6 (middle), with overlay (right). Core activation regions persist while additional patches emerge at deeper layers.}
\vspace{-2ex}
\label{fig:llava_two_sae}
\end{figure}

We investigate how visual representations transform through LLaVA by training SAEs at two points: the projector output (which converts vision encoder features to LLM tokens) and layer 6 of the LLM. For each image, we extract patch-level activation maps where each activation corresponds to a $14\times14$ patch.

As shown in Figure~\ref{fig:llava_two_sae}, core activated regions remain consistent between layers, while layer 6 activates additional patches. This pattern, observed across multiple images, suggests visual features persist through early LLM layers rather than immediately collapsing into abstract linguistic representations. The model appears to maintain accessible visual information while recruiting additional details during language processing.

These preliminary findings suggest SAEs can provide insights into multimodal representations. While our observations are qualitative, they demonstrate the potential of SAEs for understanding how LMMs bridge vision and language, motivating future systematic investigation.

\section{Conclusion}
In this work, we presented experiments demonstrating the representation power of Sparse Autoencoders for vision models. 
We showed that SAE features are semantically meaningful and enhance interpretable understanding across vision encoders, multimodal LLMs, and diffusion models.
For vision encoders, we found SAE features are robust for out-of-distribution generalization, capturing general purpose features, as well as being able to recover hierarchical semantic structures.
In diffusion models, we demonstrated semantic steering through SAE manipulation, providing an easy to use framework for additional controllability for diffusion models.
For multimodal models, our analysis suggests the shared vision-language representations is an emergent propertity of LLMs.
These applications highlight SAEs as a unified interpretability tool across diverse vision architectures.
Our experiments demonstrate the strong potential of SAEs in improving interpretability, OOD generalization, and steerability in the visual domain, providing a foundation for future application of SAEs in vision models.

{\small
\bibliographystyle{ieeenat_fullname}
\bibliography{egbib}
}

\clearpage
\appendix
\section{SAEs Layer-by-layer}
\label{sec:dino-by-layer}

\begin{figure}[tb]
    \includegraphics[width=\linewidth]{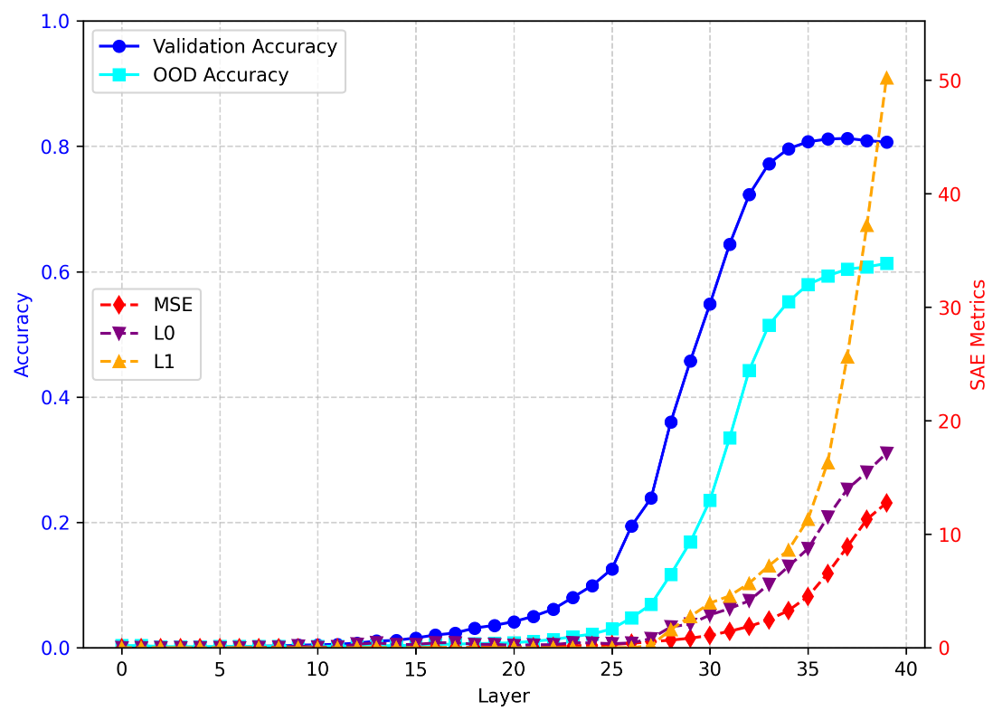}
    \caption{Results of training a ReLU SAE on every layer of DINOv2.}
    \label{fig:dino_by_layer}
\end{figure}

In figure \ref{fig:dino_by_layer} we find early layers of DINOv2's CLS representation contain no information. This suggests SAEs can be used to identify meaning in a given model's token representations simply by measuring the unsupervised SAE metrics.

\section{Steering Stable Diffusion3.5}
\label{sec:diffusion35}
\begin{figure*}[tb]
    \includegraphics[width=\linewidth]{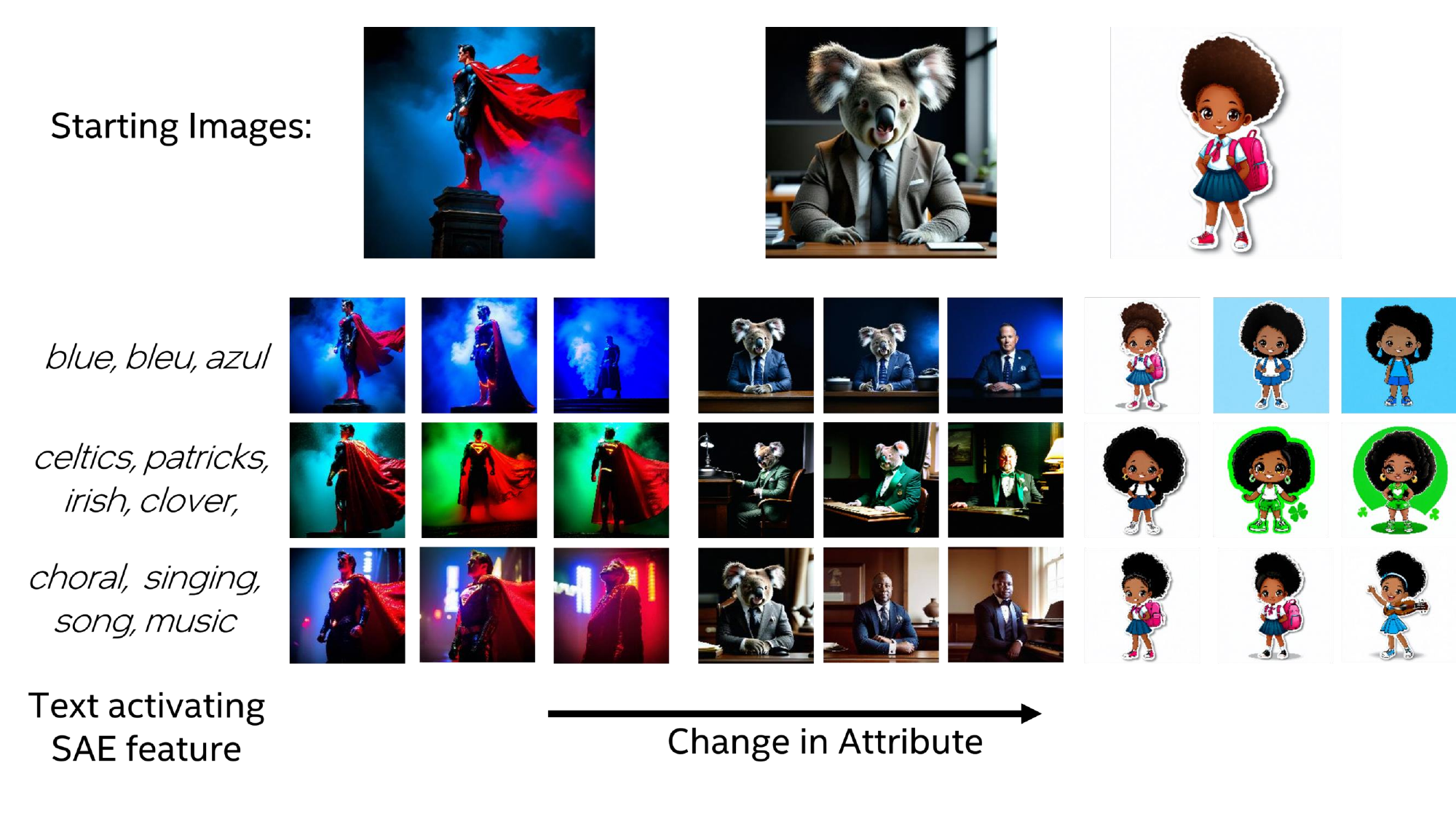}
    \caption{\textbf{Steering StableDiffusion 3.5}. We steer the generation of three starting images using three learned SAE features. While SD3.5 is challenging to steer, using an SAE to steer is technically possible.}
    \label{fig:sd3_example}
\end{figure*}

In figure \ref{fig:sd3_example}, we show examples of steering on StableDiffusion3.5~\cite{stabilityai2024sd35}. We train an SAE on the full text output of the three text encoder models, treating each positional representation as independent. We found some highly activate text embeddings, then used those to steer the model at test time. While these activations are consistent, many of the learned SAE activations were not semantically meaningful. The exact prompts for the starting images are: 
\begin{enumerate}
    \item \textit{close up shot of Supermans cape flapping in the wind, glowing neon, in the style of Yoji Shinkawa, wide shot, dark and gritty Superman film, visual striking colors, neon demon vibes, epic, Superman is standing on a tall statue, his red cape is flapping in the night sky, shot on Afga Vista 400, natural lighting}
    \item \textit{A fit person wearing a suit and tie sitting at the desk. Head of a Koala taken with a Canon in hyperrealistic 4k on complete Black background.}
    \item \textit{back to school cute black girl cartoon sticker}
\end{enumerate}

\section{Full SAE results.}
\label{sec:sae_results}

\begin{figure*}[tb]
    \includegraphics[width=\linewidth]{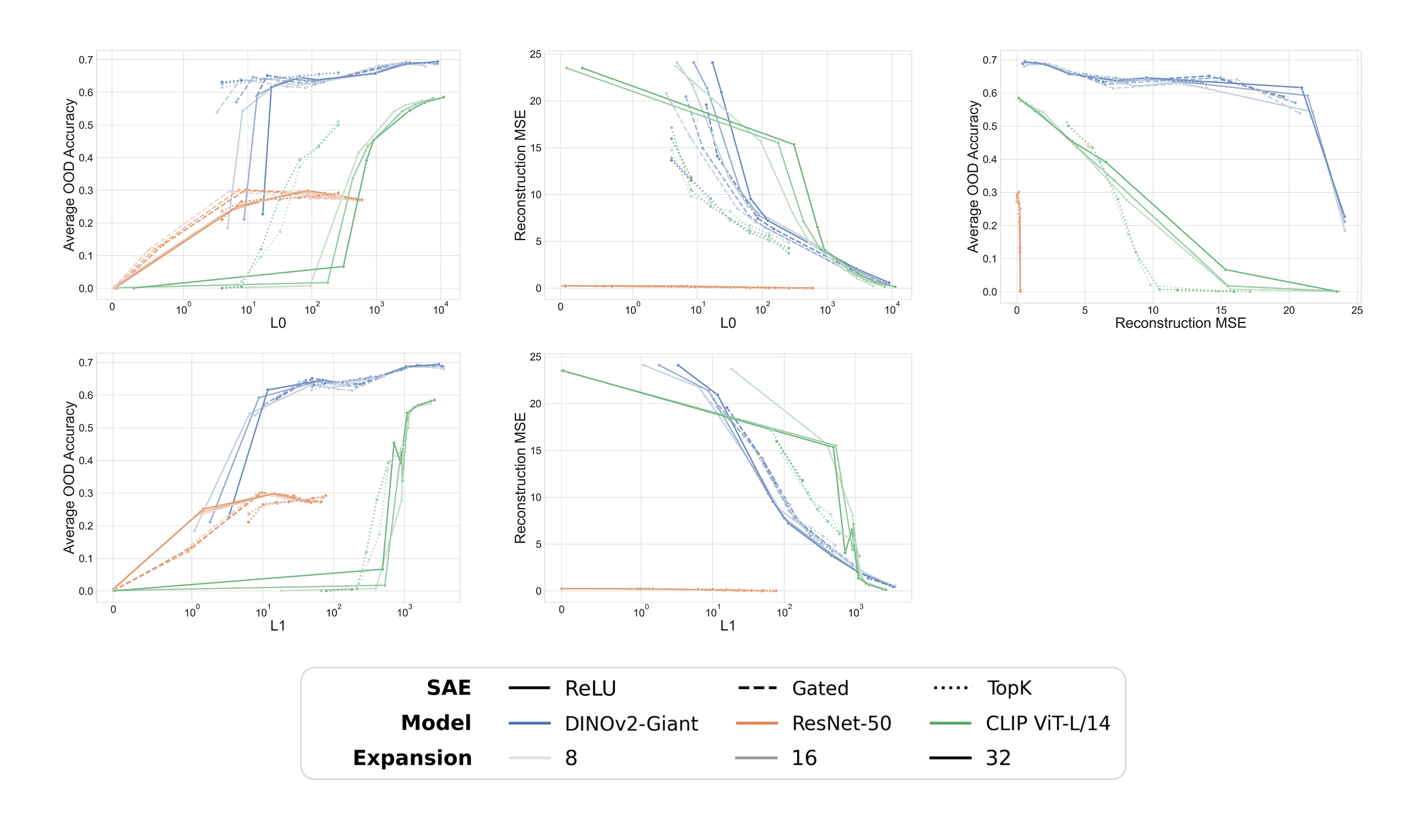}
    \caption{An expanded version of Table 1. We show OOD accuracy and Reconstruction error versus sparsity metrics.}
    \label{fig:full_sae_ood_data}
\end{figure*}

\begin{figure*}[tb]
    \includegraphics[width=\linewidth]{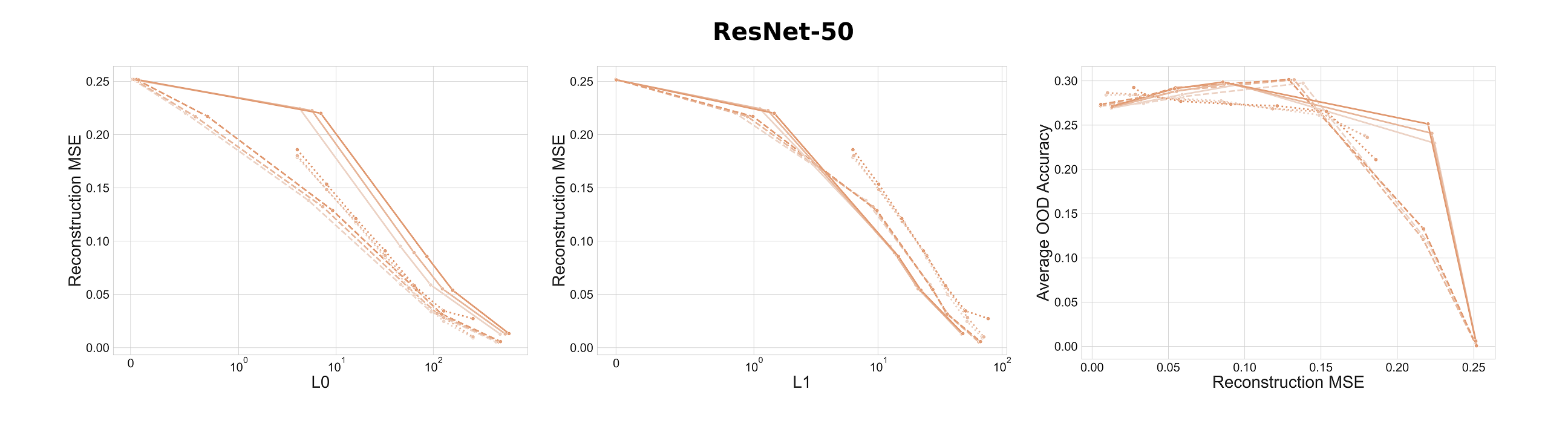}
    \caption{A detailed analysis of SAEs trained on ResNet. Across different hyper-parameters and SAE types, we find very consistent trends.}
    \label{fig:resnet_50}
\end{figure*}

\begin{figure*}[tb]
    \includegraphics[width=\linewidth]{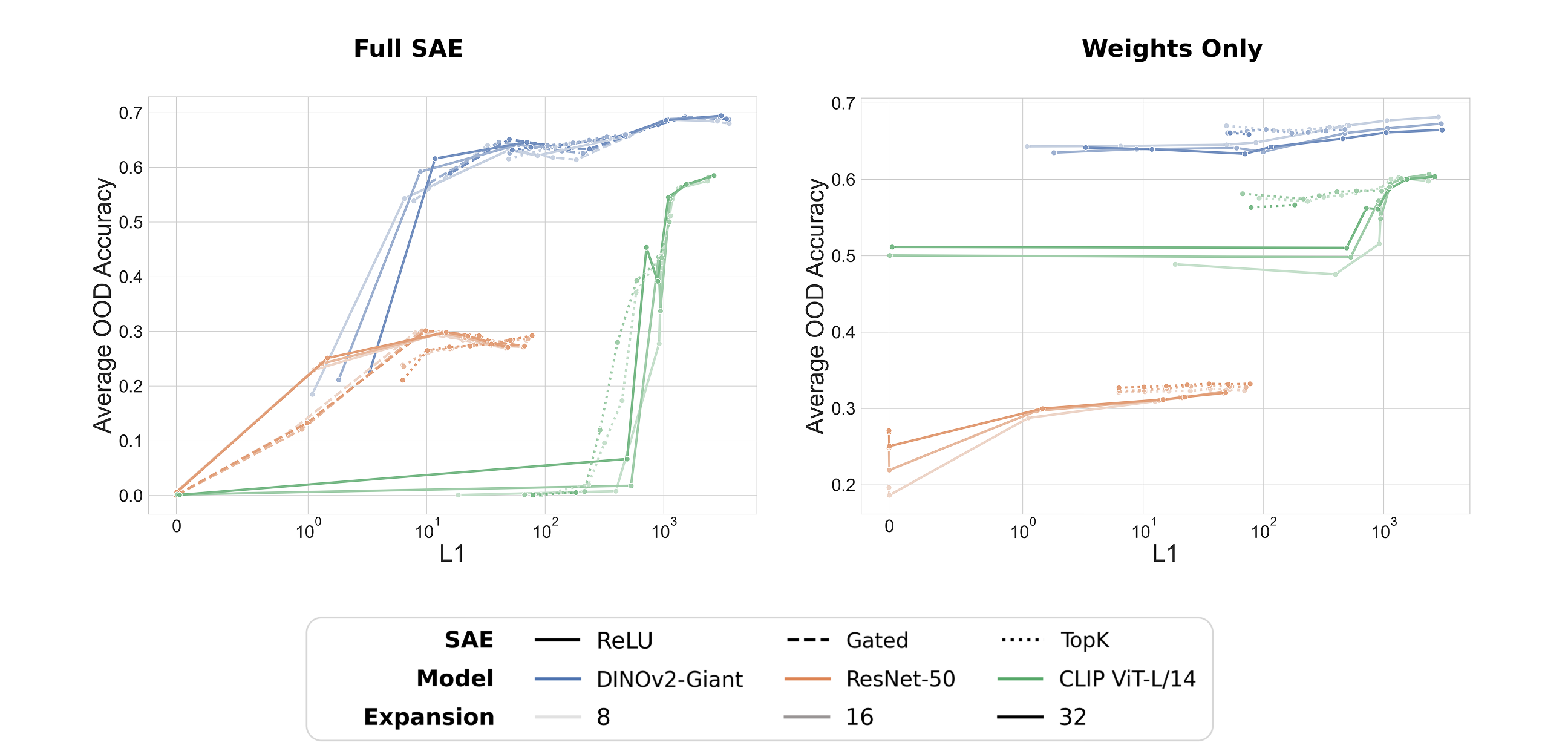}
    \caption{An experiment to measure how ignoring the SAE activations effect downstream performance on fitting a linear layer to classify ImageNet~\cite{imagenet_dataset}. We find ignoring the activations improves performance for models with original high sparsity.}
    \label{fig:sae_vs_weight}
\end{figure*}

In figure \ref{fig:full_sae_ood_data} we show detailed results for the full hyper-parameter sweep for different SAEs, Vision Models, and Expansion sizes. 
In figure \ref{fig:resnet_50} we show a detailed analysis of just ResNet~\cite{he2016deep}, finding extreme consistency across models and expansion sizes.
In figure \ref{fig:sae_vs_weight} we conduct an experiment where we fit skip the activation function for the SAE when fitting a linear layer for classification. By ignoring the activation function, SAEs trained with stricter $L_1$ penalties are able to achive high accuracy.

\label{sec:full-sae-ood-data}

\end{document}